\documentclass[a4paper,11pt]{article}
\usepackage{fullpage}
\usepackage{covington} 
\usepackage{natbib}
\usepackage{amsmath}
\usepackage{multirow}
\usepackage{rotating}

\usepackage{amsmath}
\usepackage{multirow}
\usepackage{rotating}
\usepackage[colorinlistoftodos]{todonotes}
\usepackage{linguex}

\usepackage[font=small,format=plain,labelfont=bf,up,textfont=up]{caption}
\title{Linguistic Features of Genre and Method Variation in Translation: A Computational Perspective} 

\author{Ekaterina Lapshinova-Koltunski\textsuperscript{1} and Marcos Zampieri\textsuperscript{2}\\ 
[0.5cm] \textsuperscript{1}Saarland University, Germany \\ 
\textsuperscript{2}University of Wolverhampton, United Kingdom \\
	}

\date{}

\begin{document}
\maketitle 
\thispagestyle{empty}

\begin{center}

\section*{Abstract}

\end{center}

\noindent In this paper we describe the use of text classification methods to investigate genre and method variation in an English - German translation corpus. For this purpose we use linguistically motivated features representing texts using a combination of part-of-speech tags arranged in bigrams, trigrams, and 4-grams. The classification method used in this paper is a Bayesian classifier with Laplace smoothing. We use the output of the classifiers to carry out an extensive feature analysis on the main difference between genres and methods of translation.

\vspace{5mm}

\section{Introduction}

In the present study, we use text classification techniques to explore variation in translation. We analyse the interplay between two dimensions influencing this variation: translation methods (human and machine translation) and text registers (e.g. fiction, political speeches, etc.). Our starting assumption is that the interplay between these dimensions is reflected in the lexico-grammar of translated texts, i.e. in their linguistic features. Our assumption here is that registers and methods represent two dimensions influencing linguistic properties of translations, and can thus be confounding in a specific task. For instance, if we want to automatically distinguish between human and machine translation, we need to exclude features which are rather register-specific as they can compromise the results of the classification.

In our previous work, see e.g. \cite{lapshinovaforthcomingSLE}, we used a set of features derived from theoretical frameworks, such as genre / register theory, e.g. \citet{HallidayHasan1989,Biber1995,Neumann2013}, or translationese studies, e.g. \citet{baker1993,baroni2006,VolanskyEtAl2011}. In the present analysis, we use a data-driven approach, which will help us to discover new language structures reflecting variation in translation. Classification techniques will help us to identify discriminative features of the two variation dimensions under analysis. For this, we train classifiers to distinguish translated texts according to either their register or method of translation, using the VARTRA corpus \citep{lapshinova2013BUCC}, a collection of English to German translations. Our assumption is that text classification methods can level out discriminative features of different translation varieties that intuition alone cannot grasp; thus enabling us to investigate in more detail the properties of each of them. More than the classification results {\em per se}, we use level out interesting linguistic features that can be further used in linguistic analysis and NLP applications.

Text classification methods have been applied in a wide range of tasks such as spam detection \citep{medlock08}, native language identification (NLI) \citep{gebreetal2013} temporal text classification \citep{niculae14}, and the identification of lexical complexity in text \citep{malmasi2016LTG}. In the aforementioned studies, researchers are interested in how well classification methods can perform or, in other words, how reliably these methods are able to attribute correct labels to a set of texts. Therefore, most researchers in text classification are concerned in exploring features and algorithms that deliver the best performance for each task. In recent works \cite{diwersy14,zampierietal13}, however, text classification methods were proposed  to investigate language variation across corpora (e.g. diatopic and dialectal variation) using linguistically motivated features. 

In this paper, we propose an approach to automatically classify translated texts regarding register and method of translation. We are interested not only in obtaining state-of-the-art classification performance, but also in leveling out interesting linguistic features from the data. The features used here constitute combinations of part-of-speech (POS) tags. These POS combinations represent, however, language patterns, e.g. a finite auxiliary verb followed by a participle represents a verbal phrase in a passive voice. Analyzing sets of the POS combinations that result from the classification experiments, we try  identify those that are specific for the classes under analysis.

This paper is structured as follows: the following section presents the theoretical background, as well as the related work. Here, we also describe our previous experiments on text classification for the analysis of translation variation. In Section \ref{sec:methods}, we introduce the dataset, as well as the methodology applied for the analysis. Section \ref{Sec:Results} presents the results of the classification. We further investigate the results in Section \ref{Sec:MIF}, in which we concentrate on the analysis of features specific for translation methods and genres. Section \ref{sec:conclusion} summarizes the findings and presents a discussion on the related issues.

\section{Theoretical Background and Related Work}\label{sec:TheoryRelated}
\subsection{Theoretical Background}\label{sec:Theory}
Translation is influenced by several factors, including the source and the target language, registers or genres a text belongs to, as well as the translation method involved. Since the present study focuses on genre and method variation, we will also base our research on the studies related to this type of variation.

Genre-specific variation of translation is related to studies within register and genre theory, e.g. \cite{HallidayHasan1989}, \cite{Biber1995}, which analyse contextual variation of languages. In the present paper, we use  the term {\bf genre} and not {\bf register}, although they represent two different points of view covering the same ground, see e.g. \cite{Lee2001}, and we use the latter in our previous studies, see e.g. \cite{lapshinovaforthcomingSLE} and \cite{LapshinovaVela2015Disco}. Mostly, we refer to genre when speaking about a text as a member of a cultural category, about a register when we view a text as language. However, in this study we consider both as lexico-grammatical characterisations, conventionalisation and functional configuration determined by a context use. 

The differences between genres can be identified through a corpus-based analysis of phonological, lexico-grammatical and textual (cohesion) features in these genres; see the  studies on linguistic variation by~\citet{Biber1995} or \citet{BiberEtal1999}, and linguistic variation among genres can be traced in the distribution of these features. 

Multilingual studies concern linguistic variation across languages, comparing genre and register settings specific for the languages under analysis, e.g.~\citet{Biber1995} on English, Nukulaelae Tuvaluan, Korean and Somali, and ~\citet{Croco2012} and \citet{Neumann2013} on English and German. The latter two also consider this type of linguistic variation in translations. Other translation scholars e.g.~\citet{Steiner2004} and~\citet{House2014}, also pay attention to genre and register variation when analysing language in a multilingual context of translation. However, they either do not account for the  distributions of the corresponding features,
or analyse individual texts only. In the works by \citet{DeSutterEtAl2012} and \citet{DelaereDeSutter2013}, register-related differences are also described for translated texts. Yet, these differences are identified on the level of lexical features only.

The features that are most frequently used in studies on variation in corpus-based approaches are of shallow character and include lexical density (LD), type-token-ratio (TTR), and part-of-speech (POS) proportionality. \citet{Steiner2012} uses these features to characterise profiles of various subcorpora distinguished by language (English and German), text production type (translation and original) and eight different registers. The author defines a number of contrast types including register controlled ones which implies (1) contrasts within one register between English and German, and (2) contrasts between registers within each of the languages, see \citep[p. 72]{Steiner2012}. In our analysis, we consider genre variation only within translations.

Applying a quantitative approach, \citet{Neumann2013} analyses an extensive set of linguistic patterns reflecting register variation and shows the differences between the two languages under analysis. The author also demonstrates to what degree translations are adapted to the requirements of different registers,  
showing how both register and language typology are at work. 

\citet{SleGecco} show that register variation is also relevant for a number of textual phenomena. They analyse structural and functional subtypes of coreference, substitution, discourse connectives and ellipsis on a dataset of several registers in English and German. They are able to identify contrasts and commonalities across the two languages and registers with respect to the subtypes of all textual phenomena under analysis. The authors show that these languages differ as to the degree of variation between individual registers in the realisation of the phenomena under analysis, i.e. there is more variation in German than English. They attest the main differences in terms of preferred meaning relations: a preference for explicitly realising logico-semantic relations by discourse markers and a tendency to realise relations of identity by coreference. Interestingly, similar meaning relations are realised by different subtypes of discourse phenomena in different languages and registers.

Whereas attention is paid to genre settings in human translation analysis, they have not yet been considered much in machine translation. There exist some studies in the area of statistical machine translation (SMT) evaluation, e.g. errors in translation of new domains \cite{IrvineEtAl2013}. However, the error types concern the lexical level only, as the authors operate solely with the notion of domain and not genre
. Domains represent only one of the genre parameters and reflect what a text is about, i.e.~its topic, and further settings are thus ignored. 
Although some NLP studies, e.g.~those employing web resources, do argue for the importance of genre conventions, see e.g. \citet{Santini2010}, genre remains out of the focus of machine translation. In the studies on adding in-domain bilingual data to the training material of SMT systems \citep{Wu2008} or on application of in-domain comparable corpora \citep{IrvineCallison-Burch2014}, again, only the notion of domain is taken into consideration. 

Variation in terms of translation method has not received much attention so far. There are numerous studies in the context of NLP that address both human and machine translations \citep{Papineni2002,BabychHartley2004}. Yet they all serve the task of automatic MT system evaluation and focus solely on translation error analysis, using human translation as a reference in the evaluation of machine translation outputs. Evaluations serves the task to prove to what extent automatically translated texts (hypothesis translations) comply with the manually translated ones (reference translations). The ranking of machine-translated texts is based on scores produced with various metrics. The metrics applied in the state-of-the-art MT evaluation are automatic and language-independent: BLEU and NIST~\citep{NIST2002}. However, since they do not incorporate any linguistic features, BLEU scores need to be treated carefully, which was demonstrated by~\citet{BurchEtAl2006}. This fact has been advancing the development of new automatic metrics, such as METEOR~\citep{Meteor2014}, Asiya~\citep{Asyia2014} and VERTa~\citep{Verta2014}. They incorporate lexical, syntactic and semantic information into their scores. 
The accuracy of the evaluation methods is usually proven through human evaluation. More specifically, the automatically provided scores are correlated with the human judgements which are realised by ranking MT outputs~\citep{wmt2014,VelaGenabith2015} and others. 
Some of the existing metrics incorporate linguistic knowledge. 

There are even more works on MT evaluation that operate with linguistically-motivated categories, e.g.~\citet{PopovicNey2011} or~\citet{FishelEtAl2012}. However, none of them provides a comprehensive analysis of the differences between human and machine translation in terms of specific linguistically motivated features.
In fact, the knowledge on the discriminative features of human and machine translation can be derived from the studies operating with machine learning procedures for MT evaluation, such as \citet{Beer2014} or \citet{GuptaEtAl2015}. \citet{CorstonEtAl2001} use classifiers that learn to distinguish human translations from machine ones. These classifiers are trained with various features including lexicalised trigram perplexity, part of speech trigram perplexity and linguistic features such as branching properties of the parse, function word density, constituent length, and others. Their best results are achieved if perplexity calculations were combined with finer-grained linguistic features. Their most discriminatory features that differentiate between human and machine translations are not just word n-grams. They include the distance between pronouns, the number of second person pronoun, the number of function words, and the distance between prepositions. 

\citet{VolanskyEtAl2011} operate with translationese-inspired features, and are able to distinguish between manual and automatic translations in their dataset with 100\% accuracy. However, the manual and automatic translations they are using have different source texts. We believe that the distinction they are able to achieve is not the distinction between translation methods, but rather between different underlying texts, since their most discriminatory features are the ones that show good performance in any text classification task (token n-grams). \citet{ElHajEtAl2014} make use of readability as a proxy for style and analyse consistency in translation style considering how readability varies both within and between translations. They compare Arabic and English human and machine translations of the originally French novel ``The Stranger'' (French: L'\'{E}tranger). The results show that translations by humans (both male and female) are closer to each other than to automatic translations. The authors also measure closeness of translations to the original in terms of the selected measures, which should serve as an indicator of translation quality.

To the best of our knowledge, there have not been many studies published about the interplay between the two dimensions influencing translation that are in focus of our study. \cite{KrugerRooy2012} try to answer the question on the relationship between register and the features of translated language. Their hypothesis was that the translation-related features would not be strongly linked to register variation suggesting that in translated text reveal less register variation, or sensitivity to register, which is a consequence of translation-specific effects. However, their findings provide limited support for this hypothesis. They state that the distribution and prevalence of linguistic realisations of the features of translated language may vary according to register. Therefore, the concept of translated language should be more carefully analysed and defined in terms of registers \citep[p. 61--62]{KrugerRooy2012}. \citet{JensetMcGillivray2012} analyse the interaction between registers, source language and translators' background on the basis of morphological features. The interaction between the dimensions of register, author and translator was also analysed by \citet{JensetHareide2013} who use patterns of sentence alignment as features. 

Thus, there is no comprehensive description of the linguistic features that represent the dimensions of translation variation. We analyse the interplay between the dimensions of genre (register) and method trying to detect specific features that reflect this interplay.

\subsection{Previous Experiments}\label{sec:Previous}

In our previous analyses \citep{ZampieriLapshinova2015}, we applied text classification methods on a set of English-German translations. We used two different sets of features: n-grams taking 1) all word forms into account and 2) semi-delexicalized text representations -- all the nouns were replaced with placeholders, which represented the novelty of that approach. Our task was two-fold: (a) to discriminate between different genres (fiction, political essays, etc., a total of seven classes); and (b) to discriminate between translation methods (human professional, human student, rule-based machine and two statistical systems -- five classes).

We performed several classification tasks: (1) We use word n-grams to train five translation method and seven genre classes; (2) We use delexicalised n-grams to classify four and five translation method classes and seven register classes; (3) We use delexicalised n-grams to classify between human and machine translation. 

The results of the first experiment show that the classifier performs better for the distinction of genres than of translation methods (F-measure of 57.30\% and 35.30\% respectively). This result is not surprising, as content words (including proper nouns) are domain specific and, therefore, the classifier can better differentiate between genres that vary in their domains. The results of this experiment shows that it is important to use (semi-)delexicalised features in a dataset that represents both dimensions of variation in translation -- genre and method. 

The results of the second experiment show that delexicalised features reduce the performance of the genre classifier (from 57.30\% to 45.40\%) but increase the performance of the translation method classifier (from 35.30\% to 43.10\%), especially if we reduce the dataset to four classes instead of five (we concatenate both statistical machine translation outputs). The results of this experiment confirm the importance of (semi-)delexicalised features, as we achieve similar scores for the analysis of both dimensions of variation in our data.

In the last experiment, we reduce the number of translation method classes to two -- human and machine. This classification is less fine-grained and represents manual and automatic procedures of translation. As expected, this experiment delivers better classifier results: up to 60.5\% F-measure in distinguishing between manually and automatically translated texts.

In the last step, we performed qualitative analysis of the output features paying attention to those which turned to be most informative for the corresponding classification task. In this way, we were able to identify a set of semi-delexicalized n-grams that are discriminative for either certain genres or translation methods in our data. This step was manual and included evaluation of trigrams only, as the performance of trigram models achieved the best results in the classification task. We generated two lists of features specific either to human or machine translation, and fourteen lists of features discriminating genre pairs. The features informative for translation method included full nominal phrases that differentiated in the type of determiners (articles in human and possessives in machine translations), personal pronouns expressing coreference that differentiated in the grammatical number (singular in human and plural in machine translations), event anaphors that differentiated in the type of pronouns (demonstrative in human and personal in machine), etc. These features differed from those specific for genre identification. For instance, for the discrimination between political essays ad fictional texts, discourse markers expressing different relations turned to be informative. Moreover, the lists included also features related to verbal phrases, e.g. passive vs. active voice, infinitives and modal verbs differing in their meaning.

In this paper we base upon the results of this analysis: we use a two-member classification of translation methods (manual and automatic) and seven-member classification for genres. Moreover, we decide to fully delexicalise the features and run our experiments on delexicalised features instead of semi-delexicalised ones.

\section{Methods}
\label{sec:methods}

\subsection{Data}
\label{subsec:corpus}


For the purpose of our study we looked for suitable translation corpora containing different genres and methods of translation. The only corpus known to us that possesses these characteristics is 
VARTRA \citep{lapshinova2013BUCC}. VARTRA comprises multiple translations from English into German. These translations were produced with five different translation methods as follows: (1) human professionals (PT1), (2) human student translators (PT2), (3) a rule-based MT system (RBMT), (4) a statistical MT system trained with a large quantity of unknown data (SMT1) and (5) a statistical MT system trained with a small amount of data (SMT2). 

VARTRA contains texts from different genres, namely: political essays (ESS), fictional texts (FIC), instruction manuals (INS), popular-scientific articles (POP), letters of share-holders (SHA), prepared political speeches (SPE), and touristic leaflets (TOU). Each sub-corpus represents a translation variety, a translation setting which differs from all others in both method and genre (e.g. PT1-ESS or PT2-FIC, etc.). The corpus is tokenised, lemmatised, tagged with part-of-speech information, segmented into syntactic chunks and sentences. The annotations were obtained with Tree Tagger \citep{schmid94}.

Before classification was carried out, we split the corpus into sentences.\footnote{The decision to split the corpus into sentences was motivated by the amount of texts available in the VARTRA corpus. Splitting the corpus into sentences generated enough data points for text classification and made the task more challenging.} The length of each sentence varies between 12 and 24 tokens. This results in a dataset containing 6,200 instances. 


The features used in the experiments we report in this paper were based on the combinations of POS tags 
arranged in form of bag-of-words (BoW), bigrams, trigrams, and 4-grams.\footnote{Note that in this paper we make a clear distinction between BoW and unigrams. The BoW models used in this paper do not comprise any smoothing method, whereas the $n$-gram models are calculated using Laplace smoothing.} In Example \Next, we illustrate the representation of the sentences in the corpus. \Next[a] represents a sentence from the corpus, \Next[b] shows the representation, where all nouns are substituted with the placeholder {\em PLH} resulting in what we call a semi-delexicalized text representation. 

\ex.
\a. {\sl Die weltweiten Herausforderungen im Bereich der Energiesicherheit erfordern über einen Zeitraum von vielen Jahrzehnten nachhaltige Anstrengungen auf der ganzen Welt.}
\b. {\sl Die weltweiten PLH im PLH der PLH erfordern über einen PLH von vielen PLH nachhaltige PLH auf der ganzen PLH.}
\c. {ART ADJA NN APPRART NN ART NN VVFIN APPR ART NN APPR PIAT ADJA ADJA NN APPR ART ADJA NN.}

\noindent This type of representation lies 
between fully delexicalized representations, such as the one proposed by \cite{diwersy14} for the study of variation in translation and diatopic variation of French texts, and the fully lexicalized representation, common in most text classification experiments, which uses all words in text without any substitution. This representation minimizes topic variation. Previous studies have shown that named entities significantly influence the performance of text classification systems \citep{zampierietal13,goutteetal2016}. We used this representation in our previous experiments that we describe in Section \ref{sec:Previous} above.

In \Last[c], we use a fully delexicalized representation representing texts only using the POS annotation available at the VARTRA corpus. \cite{zampierietal13} show that classification experiments using POS and morphological information as features can not only be linguistically informative, but also achieve good performance in discriminating between texts written in different Spanish varieties. Therefore, we use this representation to test whether this is also true for translated texts. 

The underlying tagset used in TreeTagger is ``Stuttgart/Tübinger Tagsets'' (STTS)\footnote{http://www.ims.uni-stuttgart.de/forschung/ressourcen/lexika/TagSets/stts-table.html}, one of the commonly used tagsets for German. In Table \ref{tab:tagset}, we illustrate a segment from the corpus with an explanation of selected tags.

\begin{table}[ht!]
\caption{Illustration of the STTS-tagged corpus segment}
\centering
\begin{tabular}{lll}
\hline
\bf word &\bf POS &\bf category \\
\hline
Die	&ART&	article\\
weltweiten &	ADJA	&adjective\\
Herausforderungen&	NN	&common noun\\
im	&APPRART	&preposition+article\\
Bereich	&NN &common noun\\
der	&ART&article\\
Energiesicherheit	&NN &common noun\\
erfordern	&VVFIN	& full finte verb\\
\"uber	&APPR	& preposition\\
einen	&ART&	article\\
Zeitraum	&NN	&common noun\\
\hline
\end{tabular}

\label{tab:tagset}
\end{table}

The decision to use these features was motivated by our goal of investigating translation variation influenced by both genre and method, and our aim to obtain a classification method that could perform well on different corpora by capturing structural differences between these translation varieties.



\subsection{Algorithm}
\label{subsec:algorithms}

In our experiments we use a Bayesian learning algorithm similar to Naive Bayes entitled Likelihood Estimation (LE) and previously used for used for language identification in \cite{zampieriandgebre12,zampieri2014varclass}. Just like Naive Bayes classifiers, LE works based on an independence assumption that the presence of a particular feature of a class is not related to the presence of any other feature. The independence assumption makes the algorithm extremely fast and a good fit for text classification tasks. Bayesian classifiers are inspired by Bayes theorem represented by the following equation:

\begin{equation}\label{Bayes}
P(A|B) = \frac{P(B|A)P(A)}{P(B)}
\end{equation}

\vspace{2mm}

\noindent Where $P(A|B)$ is a conditional probability of $A$ given $B$. Using the notation by \cite{kibriya04}, a Naive Bayes classifier computes class probabilities for a given document and a set of classes $C$. It assigns each document $t_i$ to the class with the highest probability $P(c|t_i)$.

\begin{equation}\label{MNB}
P(c|t_i) = \frac{P(t_i|c)P(c)}{P(t_i)}
\end{equation}

\vspace{2mm}

\noindent LE calculates a likelihood function on smoothed n-gram language models. Smoothing is carried out using the Laplace smoothing calculated as follows:

\begin{equation}\label{Laplace}
P_{lap}(w_1...w_n)=\frac{C(w_1...w_n)+1}{N + B}
\end{equation}

\vspace{2mm}

\noindent The language models can contain characters and words (e.g. bigrams and trigrams), linguistically motivated features such as parts-of-speech (POS) or morphological categories such as the one used in \cite{zampierietal13} for the study of diatopic variation. In this paper LE is used with POS tags as features.

Models are first calculated for each particular class in the dataset. Subsequently LE calculates the probability of a document belonging to a given class. In our case classes are represented either by genres or method of translation. The function that calculates the probability of a document given a class, represented by $L$ (language model) is the following:

\begin{equation}\label{Log2}
P(L|text)= \operatorname*{arg\,max}_L \sum_{i=1}^{N} \log P(n_{i}|L) + \log P(L)
\end{equation}

\vspace{4mm}

\noindent Where $N$ is the number of $n$-grams in the test text. The language model $L$ with the highest probability determines the predicted class of each document.





\section{Classification Results}
\label{Sec:Results}

In this section, we present the results obtained in various classification experiments using a Bayesian classifier. To evaluate the performance of the classifiers we used standard metrics in text classification such as precision, recall, f-measure, and accuracy. The linguistic analysis and discussion of the most important differences between both method and genre variation will be presented later in Section \ref{Sec:MIF}.

\subsection{Translation Methods: Human vs. Machine}
\label{Sec:Method}

In this first experiment we investigate differences between translation methods. The VARTRA corpus divides translation methods into five categories, three representing automatic methods and two containing translations produced by humans. In  \cite{ZampieriLapshinova2015} we trained a classifier to discriminate between these five methods and we observed that variation was more prominent when comparing human vs. machine translations. For this reason we unify PT1 and PT2 into one class and RBMT, SMT1, and SMT2 into the other.

We represent texts using the POS tags as features as presented earlier in this chapter. We use a total of 600 texts for each class split in 400 documents for training and 200 for testing. Results are presented in terms of precision, recall, and f-measure in Table \ref{table:HUMT}. The baseline is 50\% accuracy.

\begin{table}[ht!]
	\caption{N-grams: Human x Machine}
	\centering
	\begin{tabular}{l c c c}
	\hline
		 \bf Features & \bf Precision & \bf Recall & \bf F-Measure \\ \hline	
		bigrams &  60.70\%  &  60.51\%   &  60.61\%   \\
		trigrams &  62.50\%  &   62.50\%   &  62.50\%   \\
		4-grams &  57.84\%  &   57.25\%   &  57.54\%    \\
		\hline
	\end{tabular}
\label{table:HUMT}
\end{table}

\noindent In all three settings, the model performs above the expected baseline of 50.0\% f-measure. The best performance is obtained using a POS trigram model (62.5\% f-measure and precision).


\subsection{Genres}
\label{Sec:Genre}

In this section, we train a model to automatically distinguish between seven different genres represented in our dataset. For the sake of clarity we list here the genres contained in the VARTRA corpus:  political essays (ESS), fictional texts (FIC), instruction manuals (INS), popular-scientific articles (POP), letters of share-holders (SHA), prepared political speeches (SPE), and touristic leaflets (TOU). All experiments in this section are binary classification settings in which the classifier is trained to discriminate between two genres at a time. The baseline four each setting is therefore 50\% accuracy. 

We again use POS tags as features as described in \ref{Sec:Method} arranged in bigrams, trigrams, and 4-grams. We use a total of 500 texts for each class split in 300 documents for training and 200 for testing. We evaluate the performance of our method in terms of accuracy and present results in Tables \ref{table:bigrams}, \ref{table:trigrams}, and \ref{table:fourgrams}.

In Table \ref{table:bigrams} using POS bigrams we observed that the best results were obtained when discriminating instruction manuals from fictional texts, 81.25\% accuracy. The worst results were obtained between speech and essays, 61.25\% accuracy.

 Corroborating with the findings of the previous section, we observed that for genres the overall best results are obtained when using POS trigrams. The model is able to discriminate between tourism leaflets and fictional texts with impressive results of 84\% accuracy.  


\begin{table}[ht!]
	\caption{Genres Classification in Translation in Binary Settings: POS Bigrams}
	\centering
	\begin{tabular}{l c c c c c c c}
	\hline
	 \bf Classes & \bf ESS & \bf FIC & \bf INS & \bf POP & \bf SHA & \bf SPE & \bf TOU \\ \hline	
	\bf ESS &  -  						& 78.00\%  &  75.75\% & 65.00\% & 66.25\% 	& 61.25\% & 71.75\% \\
	\bf FIC &  -  						&  -  	  &  81.25\%  & 79.50\% & 77.75\%  & 74.50\% & 80.50\% \\
	\bf	INS &  -  						&  -  & -   		 & 74.50\% & 75.50\%  & 79.00\% & 74.25\% \\
	\bf	POP &  - 						&  -  		&  -      & -     & 68.25\%  & 67.50\% & 69.00\% \\
	\bf	SHA &  -  						&  -        &  -      & -	  & -       & 66.00\% & 69.25\% \\
	\bf	SPE &  -  &  -  &  -   & - & - & - & 72.75\% \\
	\hline
	\end{tabular}
\label{table:bigrams}
\end{table}


\noindent We observe that in the vast majority of settings presented in Table~\ref{table:trigrams}, results obtained using POS trigrams were higher than those using POS bigrams.

\begin{table}[ht!]
	\caption{Genres Classification in Translation in Binary Settings: POS Trigrams}
	\centering
	\begin{tabular}{l c c c c c c c}
	\hline
	 \bf Classes & \bf ESS & \bf FIC & \bf INS & \bf POP & \bf SHA & \bf SPE & \bf TOU \\ \hline	
	\bf ESS &  -  						& 76.25\%  &  74.00\% & 66.25\% & 71.00\% 	& 61.25\% & 72.50\% \\
	\bf FIC &  -  						&  -  	  &  76.50\%  & 76.50\% & 77.75\%  & 76.25\% & 84.00\% \\
	\bf	INS &  -  						&  -  & -   		 & 74.25\% & 76.75\%  & 76.75\% & 74.75\% \\
	\bf	POP &  - 						&  -  		&  -      & -     & 71.00\%  & 67.75\% & 71.75\% \\
	\bf	SHA &  -  						&  -        &  -      & -	  & -       & 65.00\% & 68.00\% \\
	\bf	SPE &  -  						&  -  		&  -   & - & - & - 					 & 71.25\% \\
	\hline
	\end{tabular}
	
\label{table:trigrams}
\end{table}

\noindent Finally, in our last setting using POS 4-grams, we observed that this set of features do not achieve the best results in distinguishing between genres. For this reason, we preset feature analysis on POS trigrams which were the features that obtained the best results in this section.

\begin{table}[ht!]
	\caption{Genres Classification in Translation in Binary Settings: POS 4-grams}
	\centering
	\begin{tabular}{l c c c c c c c}
	\hline
 \bf Classes & \bf ESS & \bf FIC & \bf INS & \bf POP & \bf SHA & \bf SPE & \bf TOU \\ \hline	
	\bf ESS &  -  						& 73.75\%  &  75.25\% & 69.50\%  & 67.00\% 	& 65.25\% & 72.50\% \\
	\bf FIC &  -  						&  -  	  &  70.20\%  & 75.00\% & 78.25\%  & 76.00\% & 79.25\% \\
	\bf	INS &  -  						&  -  & -   		 & 68.50\% & 70.50\%  & 74.50\% & 74.50\% \\
	\bf	POP &  - 						&  -  		&  -      & -     & 71.25\%  & 68.75\% & 70.75\% \\
	\bf	SHA &  -  						&  -        &  -      & -	  & -       & 67.00\% & 66.25\% \\
	\bf	SPE &  -  						&  -  		&  -   & - & - & - 					 & 73.00\% \\
	\hline
	\end{tabular}
\label{table:fourgrams}
\end{table}

	\section{Feature Analysis}
	\label{Sec:MIF}
	
	Text classification allows us to not only measure how well certain subcorpora (e.g. human and machine translations) are distinguished from each other, but also which individual features contribute to this distinction. Therefore, we analyse the output features resulting from the classification in this section. The main aim here is to identify the most informative features from the delexicalized n-grams in our experiments and to interpret them in terms of linguistic categories. This step is manual and carried out by looking through the most informative features and thus discriminative for certain genres and translation methods in our translation data. 
	
	Delexicalized trigrams consist of a sequence of words and placeholders, e.g. (1) {\sl ART NN VMFIN} (2) {\sl . KON PPER}, etc. Intuitively, we try to recognise more categories on a more abstract level of linguistic description, i.e. category of modality expressed through modal verbs, discourse-building devices, such as discourse markers and coreference and others for the given trigrams. Thus, example (1) represents a finite clause containing a full nominal phrase, and example (2) represents a pattern related to the level of discourse: a Connector at sentence start followed by a personal pronoun that likely refers to something previously mentioned in the text. We decide for the evaluation of trigrams, as the performance of trigram models achieved the best results in both classification tasks. 
	
	\subsection{Translation Methods}\label{Sec:MIF:method}
	The classification results for the distinction of translation methods outputs two lists of features: (1) the list of the features specific for human translations and (2) the list of features specific for machine translation. We analyse up to the first 20 features per translation method, summarising our observations in Table \ref{table:M-features}. 
	
	\begin{table}[ht!]
    	\caption{Features discriminating between human and machine translations}
		\centering
		\begin{tabular}{p{1.6cm}|p{4.85cm}|p{4.85cm}}
			\hline
		&	\bf human &\bf  machine \\
			\hline
			\multirow{7}{*}{Discourse}
& conjunct at sent.start followed by a personal referring expression &  conjunct at sent.start followed by a full NP \\
&  & adverbial at clause start\\
&  &  conjunct at clause start \\
& coreference at sentence start (demonstrative) &  full and pronominal referring expressions at clause start\\
&            coreference and negation&\\

\hline
Modality &-&+\\
\hline
\multirow{5}{*}{Phrases}&
NP connected to other phrases & V2 phrases followed by a definite NP\\
&conj linking NP & conj linking VP\\
&NP with named entities & NP describing location \\ 
&                               & predicative adjectives \\ 
&                               & locative prepositional phrases \\ 
\hline
\multirow{3}{*}{Verbs}&verb subcategorisation patterns& apposition\\
&                               & V2 structure \\ 
&                               & imperative constructions \\
		\hline
		\end{tabular}
	
\label{table:M-features}
	\end{table}


As seen from the lists, both translation methods have similar types of features that differentiate them from each other: they can be classified in terms of more abstract linguistic categories, such as discourse and modality. And some of them concern the preferred typology of phrases which can be related to the style of writing: nominal vs. verbal. The differences between the features discriminating between human and machine translations are visible on a more fine-grained level. For example, if we take into account morpho-syntactic preferences of discourse phenomena, we observe differences in the position of cohesive triggers: in machine translations, several patterns contain a punctuation mark in the first position, which means that the observed pattern often represents sentence and clause start. Human translations rather show preferences for more sentence-starting devices. Example \Next reveals the possible reasons for this observation: human translators tend to split longer sentences into several ones \Next[b], whereas a machine translation system keeps the structure \Next[c] as it was in the source \Next[a].

\ex.
\a. {\sl  He used it to modernise the castle but he must have skimped on the kitchen, since 1639 it fell into the sea and carried away the cooks and all their pots.}
\b. {\sl Er erbeutete das vom Schiff mitgeführte Gold und benutzte es zur Modernisierung des Schlosses. Allerdings scheint er beim Ausbau der Küche etwas geknausert zu haben, denn dieser Teil des Schlosses rutschte im Jahre 1639 ins Meer ab. Die Köche wurden mitsamt Töpfen weggespült.}
\c. {\sl Er benutzte es, um das Schloss zu modernisieren, aber er muss auf dem Küchentisch gespart haben, seit 1639 ins Meer fiel und trugen die Köche und alle ihre Töpfe..}


Another difference that is clearly seen from the examples in the corpus data is the preference of human translations for conjuncts, whereas for machine translations, subjuncts and adverbials seem to be more typical. This is seen in the illustration in example \Next.

\ex. 
\a. 
Human: {\sl \underline{Und} er wandte sich von der goldenen Dame ab und hätte gar zu gern die silberne genommen...}
\b. Machine: {\sl Darüber hinaus muss der Bohrvorgang verfeinert, \underline{so dass} die tieferen Aquiferen nicht durch Arsen-Lager Wasser rann von den flachen Grundwasserleiter durch die Bohrungen selbst vergiftet werden.}

The multiword conjunction {\sl so dass} is very frequent in machine translations in our data: the total of 108 occurrences as compared to 6 occurrences in human translations. A closer look at the data reveals that all the occurrences of {\sl so dass} are found in machine translations  produced with a statistical system trained on a large amount of data.

Human translations have also a number of features related to modality  expressed via modal verbs. Our additional quantitative analysis of the modal verb distributions shows that, in general, human translations contain slightly more modal verbs than the machine ones: 16.49\% vs. 15.72\% out of all sentences in our data. This means that although modal verbs are more frequent in human translations, linguistic patterns with modals are more distinctive for machine-translated texts. 

There is a difference in adjectival constructions. Predicative adjectives that turn to be discriminative for machine translations are also more frequent in this translation variety than in the human one (24.86\% vs. 23.72\%).

\ex. 
\a. {\sl The roads are \underline{excellent}, with miles of motorway and dual carriageway...}
\b. {\sl Es gibt \underline{ausgezeichnete} Straßen, davon ungefähr 112 km Autobahn und noch weit mehr Kilometer mit zweispurigen Fahrbahnen.}
\c. {\sl Die Straßen sind  \underline{sehr gut}, mit Meilen von Autobahn und Schnellstraße...}

As seen from example \Last, the machine-translated sentence \Last[c] is closer to the source one \Last[a] in terms of the predicative vs. attributive usage of the adjective ({\sl are excellent -- sind sehr gut}). At the same time, human translation \Last[b] is closer to the source in terms of lexical choice ({\sl excellent -- ausgezeichnet})

Another interesting difference is the prevalence of nominal structures in human translations as opposed to machine ones (46\% vs. 24\%) in the analysed trigram patterns. At the same time, machine-translated texts in our corpus contain more verbal phrases under their discriminative features (50\% vs. 34\%). We believe that this tendency is observed due to the shining though effect \citep{Teich2003}: German-English contrastive analyses, e.g. the one by \citet{Steiner2012} show that German has a preference for nominal structures, whereas English is more verbal. So, if a similar preference is observed in English-to-German translations, this could be interpreted as a phenomenon of shining though.



\subsection{Genres}
Using the same strategy, we generate a list of features discriminating genre pairs. 
For the sake of space, we will concentrate on the analysis of two genres only: fictional texts and political speech. For the first one, we achieve the best results in the trigram classification, whereas the classification results seem to be the worst for the second.

\subsubsection{Fictional texts}
We analyse six lists of patterns that turn out to be discriminative for fiction in the six classification tasks involving fictional texts: (1) fiction vs. political essays, (2) fiction vs. instruction manuals, (3) fiction vs. popular-scientific texts, (4) fiction vs. letters-to-shareholders, (5) fiction vs. political speeches and (6) fiction vs. tourism leaflets. 

First, we sort the patterns that occur more than once in the lists. The data contains several types of patterns: (a) informative in five classification tasks, i.e. member of five lists; (b) member of four lists; (c) member of three lists; (d) member of two lists; (e) member of one list -- informative on one particular classification task, e.g. fictional texts vs. instruction manuals. The most frequent patterns 
are considered to be the most specific ones for fictional texts, as they were informative in several classification tasks, i.e. in discriminating fictional texts from several other genres. For instance, the pattern {\sl , ADV KOUS}, e.g. {\sl , so dass / , noch bevor / , auch wenn} (a discourse marker that links a subordinate clause), is informative in the first five classification tasks (fiction against essays, instructions, letters-to-shareholders, political speeches and tourism texts).

Table \ref{table:pattern-freq} illustrates the distribution of fiction-discriminative patters. The final list comprises 170 patterns. 

\begin{table}[ht!]
\caption{Feature lists discriminating between fiction and other genres}
\centering
\begin{tabular}{l|l|r}
\hline
&list membership & types \\
\hline
(a)&5 lists & 5\\
(b)&4 lists & 5\\
(c)&3 lists & 23\\
(d)&2 lists & 49\\
(e)&1 list & 88\\
\hline
total & &170\\
\hline
\end{tabular}
\label{table:pattern-freq}
\end{table}

In the following, we analyse those that occur in most tasks (5 lists) and a subset of those that occur in one classification task only (1 list). Table \ref{table:pattern-5} illustrates the five language patterns that turned to be the most informative in most classification tasks for the discrimination of fictional texts. The last column of the table provides the information on the genre, for which the result is not valid. The first three patterns are not discriminative for fictional texts, when classified against popular-scientific articles, the fourth pattern is not informative when instructional manuals are involved. And the last one is not discriminative for fictional text, when they are classified vs. letters-to-shareholders.

 \begin{table}[ht!]
\caption{Features informative for fiction in most classification tasks for fiction}
\centering
\begin{tabular}{l|l|l}
\hline
pattern & example & excluding\\
\hline
 \$, ADV KOUS & {\sl , so dass/ , noch bevor/ , auch wenn}& POP\\
 ADV KOUS PPER & {\sl so dass er/ auch wenn sie}& POP\\
 ADV VVFIN \$. & {\sl gern wiederholen ./ nebeneinander stellen}&POP\\
 \$( PPER VMFIN & {\sl (sie können/ (wir wollen/ (es dürfte}& INS\\
 PPER VMFIN PPER & {\sl sie können ihn/ wir wollen uns/ ich möchte ihm}& SHA\\
\hline
\end{tabular}
\label{table:pattern-5}
\end{table}

Most language patterns in Table \ref{table:pattern-5} are discourse-related devices, i.e. discourse markers expressing conjunctive relations ({\sl so dass}, {\sl auch wenn}) or pronouns triggering cohesive reference ({\sl er, sie, es}). The last two patterns contain also modal verbs which can be interpreted in terms of sentence or text modality.

\begin{table}[ht!]
\caption{Distribution of {\sl  \$, ADV KOUS} across genres}
\centering
\begin{tabular}{l|r|r|r|r|r|r|r}
\hline
genre & TOU & SHA & SPE & POP & ESS & INS & FIC \\
\hline
freq & 0.10  & 0.11  & 0.14  & 0.23  &  0.29  & 0.35 & 0.41 \\
\hline
\end{tabular}
\label{table:distribution-pattern}
\end{table}

The discriminative power of features does not necessarily imply a high frequency of a particular pattern in fictional texts. Nevertheless, the distribution of the first pattern across genres in our data shows that this trigram is more frequent in fiction than in the other genres, see Table \ref{table:distribution-pattern} (the number are normalised per 1000 per total number of trigrams). However, the numbers in the table do not reveal the reasons for this pattern not being discriminative in the classification task for fiction vs. popular-scientific texts.

In the last step, we analyse the list of language patterns discriminating fiction in one particular task that includes 88 trigrams. In Table \ref{table:fic-1task}, we present a summary of these patterns describing them in terms of more general linguistic categories, e.g. specific phrases or functions.

\begin{table}[ht!]
\caption{Features informative for fiction in one classification task only}
\centering
\begin{tabular}{l|l|p{5cm}}
\hline
feature & example pattern & language example \\
\hline
                        & ADJA  KON ADJA & \sl heller und dunkler / ernste und vielschichtige\\
phrases with adjectives & CARD ADJA NN & \sl zwei junge Männer / drei wunderschöne Damen \\
                        & PPOSAT NN ADJD & \sl ihre Hörner steil / ihre Lieder schwer \\
\hline
                      & ADV VAFIN PPER & \sl so ist es / dann hast du\\
phrases with adverbs  & ADV VVFIN , & \sl anders war , / unterwegs ist ,\\
                      & ADV VVFIN PPER & \sl jetzt kriegt er / dann grunzt er\\
\hline
                          & PPER ADV VVINF & \sl sie weiter sprechen / dir nur sagen \\
 coreference via pronouns & PPER VAFIN PPER  & \sl sie hatte sie /  ich habe sie \\
                          & \$. PPER VMFIN & \sl . Sie macht / Sie beginnen \\
\hline
                         & KON PPER VMFIN & \sl Aber sie möchte / und er konnte \\
discourse markers        & KOUS PPOSAT NN &\sl dass meine Mutter / ob ihr Brief\\
                         & VAFIN \$. KON &\sl würde. Aber / hatte . Und\\
\hline
\end{tabular}
\label{table:fic-1task}
\end{table}

As seen from the table, the patterns specific for fiction include adjective and adverb modification and elements that contribute to structuring discourse in a text. 
The latter are especially specific for narrative texts, which our fictional texts belong to. These observations coincide with the results of other empirical analyses on genres, e.g. those obtained by~\citet{Neumann2013}. The author also points to personal pronouns and predicative adjectives as indicators of narration and casual style which are specific for fictional texts. 


\subsubsection{Political speeches}

We proceed with the analysis of political speeches, which turned to be the hardest genre to identify in the classification with trigrams. The same analysis steps as we used for fictional texts are applied here.

First, we summarise the patterns that occur more than once in the lists. The political speeches contain less types of patterns than the fictional texts. An informative pattern can be a member of maximum four classification tasks only (for fictional texts, we also had five). So, we have fours lists of patterns: (a) informative in four classification tasks, i.e. member of our lists; (b) member of three lists; (c) member of two lists; (d) member of one list -- informative on one particular classification task, e.g. political speeches vs. fictional texts. As in the previous case with fictional texts, we consider the most frequent patterns 
to be the most specific ones for political speeches, as they contribute to the distinction of speeches from several other genres. For instance, the pattern {\sl PTKVZ \$. ART}, e.g. {\sl vor . Das / bei . Die / weiter . Das} (sentence end followed by a sentence starting with a nominal phrase with an article), is informative in the following four classification tasks: political speeches vs. fiction, instructions, letters-to-shareholders, and tourism texts. Table \ref{table:pattern-freq-spe} illustrates the distribution of speech-discriminative patters. 

The total number of patterns is smaller than that of fictional texts (156 vs. 170 pattern types). We believe that the more distinctive features a genre has, the more distinctive it is from other genres, and thus can be easily identified with automatic classification techniques.  

\begin{table}[ht!]
\caption{Features discriminating between political speeches and other genres}
\centering
\begin{tabular}{l|l|r}
\hline
list membership & types \\
\hline
(a)&4 lists & 3\\
(b)&3 lists & 18\\
(c)&2 lists & 49\\
(d)&1 list & 86\\
\hline
total& & 156\\
\hline
\end{tabular}
\label{table:pattern-freq-spe}
\end{table}

Now we will have a closer look at the patterns that are informative in most tasks and some of those that are discriminative for political speeches in one classification task only (1 list). 

\begin{table}[ht!]
\caption{Features informative for political speeches in most classification tasks for political speeches}
\centering
\begin{tabular}{l|l|l}
\hline
pattern & example & excluding\\
\hline
 PTKVZ \$. ART & {\sl vor . Das / bei . Die / weiter . Das}& ESS, POP\\
 VVFIN PPER ADV & {\sl arbeiten wir zurzeit / auch wenn sie}& POP, TOU\\
 VVPP VAINF \$. & {\sl gern wiederholen ./ nebeneinander stellen}&ESS, POP\\
\hline
\end{tabular}
\label{table:pattern-4}
\end{table}

Table \ref{table:pattern-4} illustrates the three language patterns that turned to be the most informative for the discrimination of political speeches. The last column of the table provides the information on the two genres, for which the result is not valid. All the three patterns are not discriminative for political speeches, when classified against popular-scientific articles. The first and the last patterns are not informative, when political speeches are distinguished from political essays. And the second pattern is not discriminative in the classification against tourism texts.

In the last step, we analyse the list of language patterns discriminating political speeches in one particular task that includes 86 trigrams. In Table \ref{table:speech-1task}, we present a summary of these patterns describing them in terms of more general linguistic categories, e.g. specific phrases or functions.

\begin{table}[ht!]
\caption{Features informative for political speeches in one classification task only}
\centering
\begin{tabular}{l|l|p{5cm}}
\hline
feature & example pattern & language example \\
\hline
                        & ADJA  NN KOKOM & \sl politische Themen wie / weltweite Probleme wie \\
phrases with adjectives & ADJD APPR ART & \sl wichtig für die / möglich für die \\
                        & ADV ADJA NN & \sl sehr geehrte Mitglieder / ebenfalls sprunghafte Fortschritte \\
\hline
                      & \$, PTKZU VVINF & \sl , zu sprechen / , zu bekämpfen / , zu besprechen\\
infinitive phrases   & ADJD PTKZU VVINF & \sl schwer zu entscheiden / richtig zu verteidigen \\
                      & PTKZU VVINF KON & \sl zu übernehmen und / zu unterstützen und\\
\hline
                          & PPER PPOSAT NN & \sl wir unsere Ziele / wir unseren Feinden / ich Ihre Fragen\\
 coreference via pronouns & PPER VVINF \$,  & \sl wir prüfen , /  Sie antworten , /  wir erreichen ,\\
                          & \$, VMFIN PPER & \sl , müssen wir / , möchte ich / , können wir\\
\hline
                         & \$. ADV VAFIN & \sl . Bisher haben / . Natürlich ist\\                     
discourse markers        & \$. ADV VVFIN & \sl . Möglicherweise brauchen / : Erstens versucht \\
                         & VAFIN \$. KON &\sl würde. Aber / hatte . Und\\
\hline                         
\end{tabular}
\label{table:speech-1task}
\end{table}


As seen from the table, the patterns specific for political speeches include phrases with adjective, infinitive phrases and the elements that contribute to structuring discourse in a text. 
From the first sight, there are types of patterns that are similar to those analysed for the fictional texts. However, our qualitative analysis reveals substantial differences. The main differences is caused by the difference in the register orientation: in a narration (most of the fictional texts in the data), there is more orientation towards the content, whereas in political speeches, we observe a clear orientation of the author towards the audience. It is especially prominent in coreference-related features. Fictional texts utilise a great number of third person pronouns, whereas political speeches have much more first and second person pronouns, see example \Next.

\ex. {\sl Was passierte mit den Kindern? Wollen \underline{Sie sagen}, dass \underline{Sie } eine Milliarde Dollar ausgegeben haben und nicht wissen...} [{\sl What happened to the children? Do you mean that you spent a billion dollars and you don't know...}]

This again, coincides to what was previously observed in register/genre-related analysis \citep{BiberEtal1999,Neumann2013}.




\section{Conclusion and Discussion}
\label{sec:conclusion}

This paper is, to our knowledge, the first attempt to use text classification techniques to discriminate methods and genres in translations using fully delexicalized text representations and to identify their specific features and relevant systemic differences in a single study. We report results of up to 62.50\% f-measure in distinguishing between human and machine translations using POS trigrams and 81.25\% accuracy in discriminating between speech and essays. 

The results obtained using POS tags as features was surprisingly higher than those obtained using (semi-)delexicalized representations presented in \cite{ZampieriLapshinova2015}. This seems to indicate relevant systemic differences across genres and methods of translation that algorithms relying on (morpho)-syntactic features are able to recognize.

At the same time, the results show that it is much harder to differentiate between translation methods than between different genres, even if fully delexicalized features are used. This confirms the results by \citet{lapshinovaforthcomingSLE} which shows that if we compare the influence of the genre dimensions in translation variation is much stronger than that of translation method.


The results of our analysis can find application in both human and machine translation. In the first case, they deliver valuable knowledge on the translation product, which is influenced by the methods used in the process and the context of text production expressed by the genre. In case of machine translation, the results will provide a method to automatically identify genres in translation data thus helping to separate out-of-genre data from a training corpus. 

The resulting lists of features can also be beneficial for automatic genre classification or human vs. machine distinction tasks. The knowledge on the differences between genres that these features deliver can also help to understand main differences between texts translated by humans and with machine translation systems. This information is especially valuable for translator training. Nowadays, translator training includes courses on post-editing technologies, since the application of such technologies has increased in translation industry recently. Translators need to know where the main problems (not necessarily errors) of machine-translated texts lie and what differs them from the texts by professional translators. This knowledge increases productivity in translation process.


In our future work, we want to perform a classification task for translation method within each genres. We assume that the differences between texts that differ in translation methods can be identified better, if classification is carried out within on genre only. Moreover, this will provide us with the information on how human and machine translations differ, if one particular genre is involved.

\bibliography{bibliographygenres}

\newcommand{\noop}[1]{}
\begin{thebibliography}{}

\bibitem[Babych et~al., 2004]{BabychHartley2004}
Babych, B., Hartley, A., and Sharoff, S. (2004).
\newblock Modelling legitimate translation variation for automatic evaluation
  of mt quality.
\newblock In {\em Proceedings of LREC-2004}, volume Vol. 3.

\bibitem[Baker, 1993]{baker1993}
Baker, M. (1993).
\newblock Corpus linguistics and translation studies: Implications and
  applications.
\newblock {\em Text and technology: In honour of John Sinclair}, 233:250.

\bibitem[Baroni and Bernardini, 2006]{baroni2006}
Baroni, M. and Bernardini, S. (2006).
\newblock A new approach to the study of translationese: Machine-learning the
  difference between original and translated text.
\newblock {\em Literary and Linguistic Computing}, 21(3):259--274.

\bibitem[Biber, 1995]{Biber1995}
Biber, D. (1995).
\newblock {\em Dimensions of Register Variation. A Cross Linguistic
  Comparison}.
\newblock Cambridge University Press, Cambridge.

\bibitem[Biber et~al., 1999]{BiberEtal1999}
Biber, D., Johansson, S., Leech, G., Conrad, S., and Finegan, E. (1999).
\newblock {\em Longman Grammar of Spoken and Written English}.
\newblock Longman, Harlow.

\bibitem[Bojar et~al., 2014]{wmt2014}
Bojar, O., Buck, C., Federmann, C., Haddow, B., Koehn, P., Monz, C., Post, M.,
  and Specia, L., editors (2014).
\newblock {\em {Proceedings of the Ninth Workshop on Statistical Machine
  Translation}}.
\newblock Association for Computational Linguistics, Baltimore, Maryland, USA.

\bibitem[Callison-Burch et~al., 2006]{BurchEtAl2006}
Callison-Burch, C., Osborne, M., and Koehn, P. (2006).
\newblock {Re-Evaluation the Role of Bleu in Machine Translation Research}.
\newblock In {\em Proceedings of the 11th Conference of the European Chapter of
  the Association for Computational Linguistics}, pages 249--256.

\bibitem[Comelles and Atserias, 2014]{Verta2014}
Comelles, E. and Atserias, J. (2014).
\newblock {VERTa Participation in the WMT14 Metrics Task}.
\newblock In {\em {Proceedings of the 9th Workshop on Statistical Machine
  Translation (WMT)}}, pages 368--375, Baltimore, Maryland, USA. Association
  for Computational Linguistics.

\bibitem[Corston-Oliver et~al., 2001]{CorstonEtAl2001}
Corston-Oliver, S., Gamon, M., and Brockett, C. (2001).
\newblock A machine learning approach to the automatic evaluation of machine
  translation.
\newblock In {\em Proceedings of the 3th Annual Meeting on Association for
  Computational Linguistics}, pages 148--155.

\bibitem[{De Sutter} et~al., 2012]{DeSutterEtAl2012}
{De Sutter}, G., Delaere, I., and Plevoets, K. (2012).
\newblock Lexical lectometry in corpus-based translation studies: Combining
  profile-based correspondence analysis and logistic regression modeling.
\newblock In {\em Quantitative Methods in Corpus-based Translation Studies: a
  Practical Guide to Descriptive Translation Research}, volume~51, pages
  325--345. John Benjamins Publishing Company, Amsterdam, The Netherlands.

\bibitem[Delaere and {De Sutter}, 2013]{DelaereDeSutter2013}
Delaere, I. and {De Sutter}, G. (2013).
\newblock Applying a multidimensional, register-sensitive approach to visualize
  normalization in translated and non-translated {D}utch.
\newblock {\em Belgian Journal of Linguistics}, 27:43--60.

\bibitem[Denkowski and Lavie, 2014]{Meteor2014}
Denkowski, M. and Lavie, A. (2014).
\newblock {Meteor Universal: Language Specific Translation Evaluation for Any
  Target Language}.
\newblock In {\em Proceedings of the Workshop on Statistical Machine
  Translation (WMT)}.

\bibitem[Diwersy et~al., 2014]{diwersy14}
Diwersy, S., Evert, S., and Neumann, S. (2014).
\newblock A semi-supervised multivariate approach to the study of language
  variation.
\newblock {\em Linguistic Variation in Text and Speech, within and across
  Languages.}

\bibitem[Doddington, 2002]{NIST2002}
Doddington, G. (2002).
\newblock {Automatic Evaluation of Machine Translation Quality Using N-gram
  Co-Occurrence Statistics}.
\newblock In {\em Proceedings of the 2nd International Conference on Human
  Language Technologies (HLT)}, pages 138--145.

\bibitem[El-Haj et~al., 2014]{ElHajEtAl2014}
El-Haj, M., Rayson, P., and Hall, D. (2014).
\newblock Language independent evaluation of translation style and consistency:
  Comparing human and machine translations of camus’ novel ``the stranger''.
\newblock In Sojka, P., Hor{\'a}k, A., Kope{\^c}ek, I., and Pala, K., editors,
  {\em Proceedings of the 17th International Conference TSD 2014}, volume 8655
  of {\em Lecture Notes in Computer Science}, Brno, Czech Republic. Springer.

\bibitem[Fishel et~al., 2012]{FishelEtAl2012}
Fishel, M., Sennrich, R., Popovic, M., and Bojar, O. (2012).
\newblock Terrorcat: a translation error categorization-based mt quality
  metric.
\newblock In {\em 7th Workshop on Statistical Machine Translation}.

\bibitem[Gebre et~al., 2013]{gebreetal2013}
Gebre, B.~G., Zampieri, M., Wittenburg, P., and Heskens, T. (2013).
\newblock Improving native language identification with tf-idf weighting.
\newblock In {\em Proceedings of the 8th NAACL Workshop on Innovative Use of
  NLP for Building Educational Applications (BEA8)}, Atlanta, USA.

\bibitem[Gonz\'{a}lez et~al., 2014]{Asyia2014}
Gonz\'{a}lez, M., Barr\'{o}n-Cede\~{n}o, A., and M\`{a}rquez, L. (2014).
\newblock {IPA and STOUT: Leveraging Linguistic and Source-based Features for
  Machine Translation Evaluation}.
\newblock In {\em Proceedings of the 9th Workshop on Statistical Machine
  Translation (WMT)}, pages 394--401, Baltimore, Maryland, USA. Association for
  Computational Linguistics.

\bibitem[Goutte et~al., 2016]{goutteetal2016}
Goutte, C., L\'{e}ger, S., Malmasi, S., and Zampieri, M. (2016).
\newblock Discriminating {S}imilar {L}anguages: {E}valuations and
  {E}xplorations.
\newblock In {\em Proceedings of the International Conference on Language
  Resources and Evaluation (LREC)}, pages 1800--1807, Portoroz, Slovenia.

\bibitem[Gupta et~al., 2015]{GuptaEtAl2015}
Gupta, R., Or\u{a}san, C., and van Genabith, J. (2015).
\newblock {ReVal: A Simple and Effective Machine Translation Evaluation Metric
  Based on Recurrent Neural Networks}.
\newblock In {\em Proceedings of the 2015 Conference on Empirical Methods in
  Natural Language Processing (EMNLP)}, Lisbon, Portugal.

\bibitem[Halliday and Hasan, 1989]{HallidayHasan1989}
Halliday, M. and Hasan, R. (1989).
\newblock {\em Language, context and text: \uppercase{A}spects of language in a
  social-semiotic perspective}.
\newblock Oxford University Press, Oxford.

\bibitem[Hansen-Schirra et~al., 2012]{Croco2012}
Hansen-Schirra, S., Neumann, S., and Steiner, E. (2012).
\newblock {\em Cross-linguistic {C}orpora for the {S}tudy of {T}ranslations.
  {I}nsights from the {L}anguage {P}air {E}nglish-German}.
\newblock de {G}ruyter, Berlin, {N}ew {Y}ork.

\bibitem[House, 2014]{House2014}
House, J. (2014).
\newblock {\em Translation Quality Assessment. Past and Present}.
\newblock Routledge.

\bibitem[Irvine and Callison-Burch, 2014]{IrvineCallison-Burch2014}
Irvine, A. and Callison-Burch, C. (2014).
\newblock Using comparable corpora to adapt mt models to new domains.
\newblock In {\em Proceedings of the ACL Workshop on Statistical Machine
  Translation (WMT)}.

\bibitem[Irvine et~al., 2013]{IrvineEtAl2013}
Irvine, A., Morgan, J., Carpuat, M., III, H.~D., and Munteanu, D.~S. (2013).
\newblock Measuring machine translation errors in new domains.
\newblock {\em {TACL}}, 1:429--440.

\bibitem[Jenset and Hareide, 2013]{JensetHareide2013}
Jenset, G.~B. and Hareide, L. (2013).
\newblock A multidimensional approach to aligned sentences in translated text.
\newblock {\em Bergen Language and Linguistic Studies}, 3:195--210.

\bibitem[Jenset and McGillivray, 2012]{JensetMcGillivray2012}
Jenset, G.~B. and McGillivray, B. (2012).
\newblock Multivariate analyses of affix productivity in translated english.
\newblock In Oakes, M.~P. and Ji, M., editors, {\em Quantitative Methods in
  Corpus-Based Translation Studies}, pages 301--324. John Benjamins.

\bibitem[Kibriya et~al., 2004]{kibriya04}
Kibriya, A., Frank, E., Pfahringer, B., and Holmes, G. (2004).
\newblock Multinomial naive bayes for text categorization revisited.
\newblock In {\em Proceedings of the Australian Conference on Artificial
  Intelligence}, pages 488--499.

\bibitem[Kruger and van Rooy, 2012]{KrugerRooy2012}
Kruger, H. and van Rooy, B. (2012).
\newblock Register and the {F}eatures of {T}ranslated {L}anguage.
\newblock {\em Across Languages and Cultures}, 13(1):33--65.

\bibitem[Kunz et~al., 2017]{SleGecco}
Kunz, K., Degaetano-Ortlieb, S., Lapshinova-Koltunski, E., Menzel, K., and
  Steiner, E. (2017).
\newblock Gecco -- an empirically-based comparison of english-german cohesion.
\newblock In De~Sutter, G., Delaere, I., and Lefer, M.-A., editors, {\em New
  Ways of Analysing Translational Behaviour in Corpus-Based Translation
  Studies}. Mouton de Gruyter.
\newblock TILSM series.

\bibitem[Lapshinova-Koltunski, 2013]{lapshinova2013BUCC}
Lapshinova-Koltunski, E. (2013).
\newblock {V}{A}{R}{T}{R}{A}: {A} comparable corpus for analysis of translation
  variation.
\newblock In {\em Proceedings of the Sixth Workshop on Building and Using
  Comparable Corpora}, pages 77--86, Sofia, Bulgaria. Association for
  Computational Linguistics.

\bibitem[Lapshinova-Koltunski, 2017]{lapshinovaforthcomingSLE}
Lapshinova-Koltunski, E. (2017).
\newblock Exploratory analysis of dimensions influencing variation in
  translation: The case of text register and translation method.
\newblock In De~Sutter, G., Delaere, I., and Lefer, M.-A., editors, {\em New
  Ways of Analysing Translational Behaviour in Corpus-Based Translation
  Studies}. Mouton de Gruyter.
\newblock TILSM series.

\bibitem[Lapshinova-Koltunski and Vela, 2015]{LapshinovaVela2015Disco}
Lapshinova-Koltunski, E. and Vela, M. (2015).
\newblock Measuring ’registerness’ in human and machine translation: A text
  classification approach.
\newblock In {\em Proceedings of the Second Workshop on Discourse in Machine
  Translation}, pages 122--131, Lisbon, Portugal. Association for Computational
  Linguistics.

\bibitem[Lee, 2001]{Lee2001}
Lee, D.~Y. (2001).
\newblock Genres, registers, text types, domains and styles: clarifying the
  concepts and navigating a path through the bnc jungle.
\newblock {\em Technology}, 5:37--72.

\bibitem[Malmasi et~al., 2016]{malmasi2016LTG}
Malmasi, S., Dras, M., and Zampieri, M. (2016).
\newblock {LTG at SemEval-2016 Task 11: Complex Word Identification with
  Classifier Ensembles}.
\newblock In {\em Proceedings of SemEval}.

\bibitem[Medlock, 2008]{medlock08}
Medlock, B. (2008).
\newblock Investigating classification for natural language processing tasks.
\newblock Technical report, University of Cambridge - Computer Laboratory.

\bibitem[Neumann, 2013]{Neumann2013}
Neumann, S. (2013).
\newblock {\em Contrastive Register Variation. A Quantitative Approach to the
  Comparison of English and German}.
\newblock De Gruyter Mouton, Berlin, Boston.

\bibitem[Niculae et~al., 2014]{niculae14}
Niculae, V., Zampieri, M., Dinu, L.~P., and Ciobanu, A.~M. (2014).
\newblock Temporal text ranking and automatic dating of texts.
\newblock In {\em 14th Conference of the European Chapter of the Association
  for Computational Linguistics (EACL 2014)}. Association for Computational
  Linguistics.

\bibitem[Papineni et~al., 2002]{Papineni2002}
Papineni, K., Roukus, S., Ward, T., and Zhu, W.-J. (2002).
\newblock {B}{L}{E}{U}: a method for automatic evaluation of machine
  translation.
\newblock In {\em Proceedings of the 40th annual meeting on association for
  computational linguistics}, pages 311--318.

\bibitem[Popovic and Ney, 2011]{PopovicNey2011}
Popovic, M. and Ney, H. (2011).
\newblock Towards automatic error analysis of machine translation output.
\newblock {\em Computational Linguistics}, 37(4):657--688.

\bibitem[Santini et~al., 2010]{Santini2010}
Santini, M., Mehler, A., and Sharoff, S. (2010).
\newblock Riding the rough waves of genre on the web.
\newblock In Mehler, A., Sharoff, S., and Santini, M., editors, {\em Genres on
  the Web: Computational Models and Empirical Studies}, pages 3--30. Springer.

\bibitem[Schmid, 1994]{schmid94}
Schmid, H. (1994).
\newblock Probabilistic part-of-speech tagging using decision trees.
\newblock In {\em Proceedings of International Conference on New Methods in
  Language Processing}, Manchester, UK.

\bibitem[Stanojevi\'{c} and Sima\'{a}n, 2014]{Beer2014}
Stanojevi\'{c}, M. and Sima\'{a}n, K. (2014).
\newblock {BEER: BEtter Evaluation as Ranking}.
\newblock In {\em Proceedings of the Ninth Workshop on Statistical Machine
  Translation}, pages 414--419, Baltimore, Maryland, USA. Association for
  Computational Linguistics.

\bibitem[Steiner, 2004]{Steiner2004}
Steiner, E. (2004).
\newblock {\em Translated Texts. Properties, Variants, Evaluations}.
\newblock Peter Lang Verlag, Frankfurt/M.

\bibitem[Steiner, 2012]{Steiner2012}
Steiner, E. (2012).
\newblock A characterization of the resource based on shallow statistics.
\newblock In Hansen-Schirra, S., Neumann, S., and Steiner, E., editors, {\em
  Cross-linguistic Corpora for the Study of Translations. Insights from the
  Language Pair {E}nglish-{G}erman}. Mouton de Gruyter, Berlin, New York.

\bibitem[Teich, 2003]{Teich2003}
Teich, E. (2003).
\newblock {\em Cross-Linguistic Variation in System und Text. A Methodology for
  the Investigation of Translations and Comparable Texts}.
\newblock Mouton de Gruyter, Berlin.

\bibitem[Vela and van Genabith, 2015]{VelaGenabith2015}
Vela, M. and van Genabith, J. (2015).
\newblock {Re-assessing the WMT2013 Human Evaluation with Professional
  Translators Trainees}.
\newblock In {\em Proceedings of the 18th Annual Conference of the European
  Association for Machine Translation (EAMT)}.

\bibitem[Volansky et~al., 2011]{VolanskyEtAl2011}
Volansky, V., Ordan, N., and Wintner, S. (2011).
\newblock More human or more translated? original texts vs. human and machine
  translations.
\newblock In {\em Proceedings of the 11th Bar-Ilan Symposium on the Foundations
  of AI With ISCOL}.

\bibitem[Wu et~al., 2008]{Wu2008}
Wu, H., Wang, H., and Zong, C. (2008).
\newblock Domain adaptation for statistical machine translation with domain
  dictionary and monolingual corpora.
\newblock In Scott, D. and Uszkoreit, H., editors, {\em Proceedings of the 22nd
  International Conference on Computational Linguistics (COLING-2008)}, pages
  993--1000, Manchester, UK.

\bibitem[Zampieri and Gebre, 2012]{zampieriandgebre12}
Zampieri, M. and Gebre, B.~G. (2012).
\newblock Automatic identification of language varieties: The case of
  {P}ortuguese.
\newblock In {\em Proceedings of KONVENS2012}, pages 233--237, Vienna, Austria.

\bibitem[Zampieri and Gebre, 2014]{zampieri2014varclass}
Zampieri, M. and Gebre, B.~G. (2014).
\newblock Varclass: An open source language identification tool for language
  varieties.
\newblock In {\em Proceedings of Language Resources and Evaluation (LREC)},
  Reykjavik, Iceland.

\bibitem[Zampieri et~al., 2013]{zampierietal13}
Zampieri, M., Gebre, B.~G., and Diwersy, S. (2013).
\newblock N-gram language models and {POS} distribution for the identification
  of {S}panish varieties.
\newblock In {\em Proceedings of TALN2013}, pages 580--587, Sable d'Olonne,
  France.

\bibitem[Zampieri and Lapshinova-Koltunski, 2015]{ZampieriLapshinova2015}
Zampieri, M. and Lapshinova-Koltunski, E. (2015).
\newblock Investigating genre and method variation in translation using text
  classification.
\newblock In Sojka, P., Hor{\'{a}}k, A., Kopecek, I., and Pala, K., editors,
  {\em Text, Speech and Dialogue - 18th International Conference, {TSD} 2015,
  Plzen, Czech Republic, Proceedings}, volume 9302 of {\em Lecture Notes in
  Computer Science}, pages 41--50. Springer.

\end{thebibliography}

\end{document}